%% file: naaclhlt2018.tex
\documentclass[11pt,a4paper]{article}
\usepackage{covington}
\usepackage[hyperref]{naaclhlt2018}
\usepackage{times}
\usepackage{latexsym}

\usepackage{url}

\aclfinalcopy % Uncomment this line for the final submission
\def\aclpaperid{828} %  Enter the acl Paper ID here

\usepackage{xspace}
\usepackage{mathtools}
\usepackage{amsmath}
\usepackage{amssymb}
\usepackage{color}
\usepackage{multirow}
\usepackage{tikz}
\usepackage{pifont}
\usepackage[inline]{enumitem}% http://ctan.org/pkg/enumitem
\usetikzlibrary{positioning}
\usetikzlibrary{calc}
\usetikzlibrary{fit}
\usetikzlibrary{decorations.text}
\usetikzlibrary{shapes.misc}
\usetikzlibrary{shapes.geometric,arrows}
\usetikzlibrary{intersections}
\usepackage{colortbl}
\usepackage{mathrsfs}
\usepackage{bm}
\usepackage{booktabs}
\usepackage{pgfplots}
\usepackage{arydshln}
\usepackage{framed}

\newcommand{\eg}{e.g.\ }
\newcommand{\ie}{i.e.\ }

\newcommand{\lex}[1]{\textit{#1}\xspace}
\newcommand{\aspect}[1]{\textsc{#1}\xspace}

\newcommand{\system}[2][]{\texttt{#2#1}\xspace}

\newcommand{\secref}[1]{Section~\ref{#1}\xspace}
\newcommand{\tabref}[2][]{Table#1~\ref{#2}\xspace}
\newcommand{\figref}[2][]{Figure#1~\ref{#2}\xspace}
\newcommand{\exref}[2][]{Example#1~(\ref{#2})\xspace}
\newcommand{\equref}[2][]{Equation#1~\bracketref{#2}}
\newcommand{\bracketref}[1]{(\ref{#1})\xspace}

\newcommand{\bb}[1]{\mathbb{#1}}
\newcommand{\R}{\bb{R}}
\newcommand{\mat}[2][]{\boldsymbol{#2}_{#1}}
\newcommand{\matfwd}[2][]{\overrightarrow{\bm{#2}}_{#1}}

\renewcommand{\vec}[2][]{\boldsymbol{#2}^{#1}}
\newcommand{\vecfwd}[2][]{\overrightarrow{\bm{#2}}^{#1}}
\newcommand{\vecrev}[2][]{\overleftarrow{\bm{#2}}^{#1}}
\newcommand{\vectilde}[2][]{\tilde{\bm{#2}}^{#1}}
\newcommand{\vectildefwd}[2][]{\overrightarrow{\tilde{\bm{#2}}}^{#1}}

\newcommand{\vecringfwd}[2][]{\overrightarrow{\mathring{\bm{#2}}}^{#1}}

\newcommand{\gru}{\textrm{GRU}}
\newcommand{\grufwd}{\overrightarrow{\textrm{GRU}}}

\newcommand{\softmax}{\textrm{softmax}}
\newcommand{\xentropy}{\ensuremath{\operatorname{CrossEntropy}}\xspace}

\newcommand{\T}{\mathstrut\scriptscriptstyle\top}

\newcommand{\ent}[1][]{\system[#1]{EntNet}}

\newcommand{\babi}{\system{bAbI}}
\newcommand{\sentihood}{\system{Sentihood}}
\newcommand{\cbt}{\system{CBT}}

\newcommand{\RNum}[1]{\uppercase\expandafter{\romannumeral #1\relax}}

\title{Recurrent Entity Networks with Delayed Memory Update for\\ Targeted Aspect-based Sentiment Analysis}

\author{Fei Liu \qquad Trevor Cohn \qquad Timothy Baldwin \\
         School of Computing and Information Systems \\ The University of Melbourne \\ Victoria, Australia\\
         {\tt {fliu3@student.unimelb.edu.au}} \\
         {\tt {t.cohn@unimelb.edu.au}\,\,\, {tb@ldwin.net}}}

\date{}

\begin{document}
\maketitle
\begin{abstract}

While neural networks have been shown to achieve impressive results for sentence-level sentiment analysis, targeted aspect-based sentiment analysis (TABSA) --- extraction of fine-grained opinion polarity w.r.t.\ a pre-defined set of aspects --- remains a difficult task. Motivated by recent advances in memory-augmented models for machine reading, we propose a novel architecture, utilising external ``memory chains'' with a delayed memory update mechanism to track entities. On a TABSA task, the proposed model demonstrates substantial improvements over state-of-the-art approaches, including those using external knowledge bases.\footnote{Code available at \url{https://github.com/liufly/delayed-memory-update-entnet}.}

\end{abstract}

\input{intro}
\input{model}
\input{exp}
\input{conc}

\section*{Acknowledgments}

We thank the anonymous reviewers for their valuable feedback, and gratefully acknowledge the support of Australian Government Research Training Program Scholarship and National Computational Infrastructure (NCI Australia). This work was also supported in part by the Australian Research Council.

%\bibliography{naaclhlt2018}
\bibliography{naaclhlt2018}
\bibliographystyle{acl_natbib}

\end{document}

%% file: intro.tex
\section{Introduction}

Targeted aspect-based sentiment analysis (TABSA) is the task of identifying fine-grained opinion polarity towards a specific aspect associated with a given target. The task requires classification of opinions on different entities across a range of different attributes, with the expectation that there will be no overt opinion expressed on a given entity for many attributes. This can be seen in \exref{exp:example}, e.g., where opinions on the aspects \aspect{safety} and \aspect{price} are expressed for entity \lex{LOC1} but not entity \lex{LOC2}:\footnote{Note that in our dataset, all entity mentions have been pre-nomalised to \lex{LOC$n$}, where $n$ is an index.}
   
\begin{example}
LOC1 is your best bet for secure although expensive and LOC2 is too far.\\
\label{exp:example}
\end{example}

\begin{center}
\small
\begin{tabular}{ccc}
\toprule
Target & Aspect & Sentiment\\
\midrule
LOC1 & \aspect{safety} & positive\\
LOC1 & \aspect{price} & negative \\
LOC2 & \aspect{transit-location} & negative\\ 
\bottomrule
\end{tabular}
\end{center}
\vspace{0.3cm}

The earliest work on (T)ABSA relied heavily on feature engineering \cite{Wagner+:2014,Kiritchenko+:2014}, but more recent work based on deep learning has used models such as LSTMs to automatically learn aspect-specific word and sentence representations \cite{Tang+:2016b}.

Despite these successes, keeping track of multiple entity--aspect pairs remains a difficult task, even for an LSTM. As reported in \newcite{Saeidi+:2016}, a target-dependent biLSTM is ineffective, both in terms of aspect detection and sentiment classification, compared to a simple logistic regression model with $n$-gram features. Intuitively, we would expect that a model which better captures linguistic structure via the original word sequencing should perform better, which provides the motivation for this research.

More recently, successful works in (T)ABSA have explored the idea of leveraging external memory \cite{Tang+:2016,Chen+:2017}. Their models are largely based on memory networks \cite{Weston+:2015}, originally developed for reasoning-focused machine reading comprehension tasks. In contrast to memory networks, where each input sentence/word occupies a memory slot and is then accessed via attention independently, recent advances in machine reading suggest that processing inputs sequentially is beneficial to overall performance \cite{Seo+:2017,Mikael+:2017}.

However, successful machine reading models may not be directly applicable to TABSA due to the key difference in the granularity of inputs between the two tasks: on the Children's Book Test corpus (\cbt), for example, competitive models take as input a window of text, centred around candidate entities, with crucial information contained within that window \cite{Hill+:2015,Mikael+:2017}. In TABSA, given the fine-grained nature of the task, it is common practice for models to operate at the word- rather than chunk/sentence-level. It is not uncommon to see examples like \exref{exp:example}, where the sentence starts with \lex{LOC1}, but the negative \aspect{price} sentiment towards the entity is not expressed until much later. 
Moreover, phrases such as \lex{best bet} and \lex{although} play the role of triggers, indicating that succeeding tokens bear aspect/sentiment signal.
This key difference necessitates the ability to model the delayed activation of memory updates.

In this work, we propose a novel model architecture for TABSA, augmented with multiple ``memory chains'', and equipped with a delayed memory update mechanism, to keep track of numerous entities independently.
We evaluate the effectiveness of the proposed model over the task of TABSA, and achieve substantial improvements over a number of
baselines, including one incorporating external knowledge bases, setting a new state of the art in both sentiment classification
and aspect detection.

% Local Variables:
% mode: latex
% TeX-master: "naaclhlt2018"
% End:

%% file: model.tex
\section{Methodology}
\label{sec:model}

\paragraph{Task description.} 
In TABSA, a sentence $s$ typically consists of a sequence of words: $\{w_1, \ldots, w_i, \ldots, w_m$\} where $w_i$ denotes words interleaved with one or more targets ($t$), which we assume to be pre-identified as with \lex{LOC1} and \lex{LOC2} in \exref{exp:example}. Following \newcite{Saeidi+:2016}, we frame the task as a 3-class classification problem: given a sentence $s$, a pre-identified set of target entities $T$ and fixed set of aspects $A$, predict the sentiment polarity $y \in \{\textit{positive}, \textit{negative}, \textit{none}\}$ over the full set of target--aspect pairs $\{(t, a): t \in T, a \in A\}$. For example, (\lex{LOC1},\aspect{safety}) has gold-standard polarity \textit{positive}, while (\lex{LOC1},\aspect{transit-location}) has polarity \textit{none}.

\begin{figure}[t]
\begin{center}
\resizebox{\columnwidth}{!}{
\input{ent-small.tex}
}
\end{center}
\vspace{-0.1cm}
\caption{Illustration of our model with a single memory chain at time $i$. $\sigma$, $\phi$ and $\gru$ represent \equref[s]{equ:gate2}, (\ref{equ:phi}) and (\ref{equ:gru}), while circled nodes $L$, $C$, $\odot$ and $+$ depict the location, content terms, Hadamard product, and addition, resp.}
\label{fig:model}
\vspace{-30pt}
\end{figure}
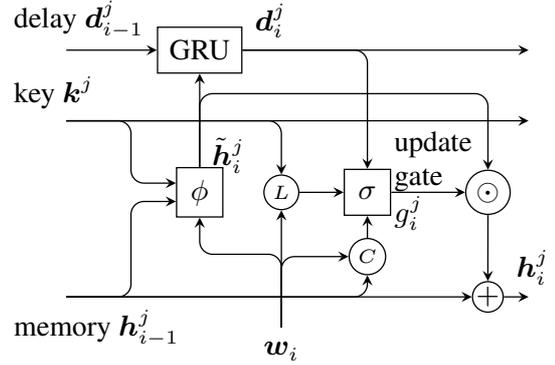

\paragraph{Proposed model.} To this end, we design a neural network architecture, capable of tracking and updating the states of entities at the right time with external memory, making it a natural fit for the task. 
Specifically, our model maintains a number of ``memory chains'' $\vec[j]{h}$, one for each entity with the key $\vec[j]{k}$ and dynamically updates the states ($\vec[j]{h}$) of them as it progresses through the sentence with the help of the delay recurrence $\vec[j]{d}$, taking previous activations into account. An illustration of our model is provided in \figref{fig:model}.

\paragraph{Delayed memory update.} Update of each memory chain is controlled by a gating mechanism, consisting of three components: the ``content'' term $\vec{w}_{i}\cdot\vec[j]{h}_{i-1}$ , the ``location'' term $\vec{w}_{i}\cdot\vec[j]{k}$ and the ``delay'' term $\vec{v}\cdot\vec[j]{d}_i$ where $\vec[j]{d}_i$ carries knowledge regarding previous activation of the gate and $\vec{v}$ is a trainable parameter vector. All three terms may lead to the activation of $g_{i}^{j}$, but differ in how they turn the gate on. While the ``location'' term causes the gate to open for memory chains whose keys ($\vec[j]{k}$) match the input, the ``content'' term triggers the activation when the content of the entities ($\vec[j]{h}_{i-1}$) matches the input. The delay term models how and when the gate was turned on in the past with a $\gru$ \cite{Chung+:2014} and how past activations should influence the current one.

More formally, with arrows denoting processing direction, the update gate is defined as:
\begin{align}
\overrightarrow{g}_{i}^{j} &= \sigma(\vec{w}_{i}\cdot\vecfwd[j]{h}_{i-1} + \vec{w}_{i}\cdot\vec[j]{k} + \vecfwd{v}\cdot\vecfwd[j]{d}_{i}) \label{equ:gate2}
\end{align}
where $\overrightarrow{g}_{i}^{j}$ is the update gate value for the $j$-th memory at time $i$,\footnote{While $\overrightarrow{g}_{i}^{j}$ could instead be a vector for finer-grained control, following \newcite{Mikael+:2017}, we use a scalar for simplicity.} $\vec[j]{k}$ is the embedding for the $j$-th entity (key), $\vecfwd[j]{h}_{i-1}$ is the hidden memory representation responsible for keeping track of the state of the $j$-th entity (content), and $\sigma$ is the sigmoid activation function. The delay recurrence $\vecfwd[j]{d}_{i}$ is defined as: 
\begin{align}
\vectildefwd[j]{h}_{i} &= \phi(\matfwd{U}\vecfwd[j]{h}_{i-1} + \matfwd{V}\vec[j]{k} + \matfwd{W}\vec{w}_{i}) \label{equ:phi}\\
\vecfwd[j]{d}_i &= \grufwd(\vectildefwd[j]{h}_{i}, \vecfwd[j]{d}_{i-1}) \label{equ:gru}
\end{align}
where $\vectildefwd[j]{h}_{i}$ is the new candidate memory vector to be incorporated into the existing memory $\vecfwd[j]{h}_{i-1}$ to form the new memory $\vecfwd[j]{h}_i$, $\phi$ is the parametric ReLU activation function \cite{He+:2015}, and $\matfwd{U}$, $\matfwd{V}$ and $\matfwd{W}$ are trainable weight matrices.

Once the update gate value has been computed, the $j$-th memory is then updated according to the intensity of $\overrightarrow{g}_{i}^{j}$: 
\begin{align}
\vecringfwd[j]{h}_{i} = \vecfwd[j]{h}_{i-1} + \overrightarrow{g}_{i}^{j}\odot\vectildefwd[j]{h}_{i}
\end{align}
where $\odot$ is the Hadamard product, and $\vecringfwd[j]{h}_{i}$ is the unnormalised memory representation for the $j$-th entity.

Essentially, gate $\overrightarrow{g}_{i}^{j}$ determines how much the $j$-th memory should be updated, factoring in three elements: (1) how similar the current input $\vec{w}_i$ is to the entity being tracked ($\vec[j]{k}$); (2) how related the current input $\vec{w}_i$ is to the state of the $j$-th entity ($\vecfwd[j]{h}_{i-1}$); and (3) how past activation should influence the current one. Update of the memory of an entity is only triggered when the gate is activated.

\paragraph{Normalisation.} Following the update, the model performs a normalisation step, allowing the memory to forget: 
$\vecfwd[j]{h}_i = \vecringfwd[j]{h}_{i}/\|\vecringfwd[j]{h}_{i}\|$
where $\|\vecringfwd[j]{h}_{i}\|$ denotes the Euclidean norm of $\vecringfwd[j]{h}_{i}$. 
As all information stored in $\vecfwd[j]{h}_i$ is constrained to be of unit length, when new information $\vectildefwd[j]{h}_i$ is added to the existing memory $\vecfwd[j]{h}_{i-1}$, the cosine distance between the original and updated memory decreases, allowing the model to forget information deemed out-of-date.

\paragraph{Bi-directionality.} We apply the above steps both forward and backward over the sentence, enabling the model to capture sentiment terms appearing before and after its associated entity. The memory representation incorporating contexts from both directions is obtained by $\vec[j]{h}_i = \vecfwd[j]{h}_i + \vecrev[j]{h}_i$, with $\vecrev[j]{h}_i$ computed analogously to $\vecfwd[j]{h}_i$.

\paragraph{Final classifier.} Our model predicts the sentiment polarity $\hat{\vec{y}}$ to the given target $\vec{t}$ and aspect $\vec{a}$ embeddings by incorporating the states of all tracked entities in the form of a weighted sum $\vec{u}$:
\begin{align}
p^{j} &= \softmax\bigg((\vec[j]{k})^{\T}\mat[att]{W}\begin{bmatrix}\vec{t}\\\vec{a}\end{bmatrix}\bigg) \\
\vec{u} &= \sum_{j} p^{j}\vec[j]{h}_m
\end{align}
where $\lbrack\,\rbrack$ denotes concatenation, $m$ is sentence length, and $\mat[att]{W}$ is a trainable weight matrix.
Here, the values of both $\vec{t}$ and $\vec{a}$ take the embedding values of their corresponding words (\ie $\vec{t}$ and $\vec{a}$ are drawn from the same embedding matrix as are the input words $\vec{w}_i$).
In the case of multi-word aspect expressions (\eg \aspect{transit-location}), we take the mean of the embeddings of the constituent words.
We then transform $\vec{u}$ to get:
\begin{align}
\hat{\vec{y}} = \softmax(\mat{R}\phi(\mat{H}\vec{u} + \vec{a}))
\label{equ:classifier}
\end{align}
Training is carried out based on cross entropy loss.
\begin{equation}
\mathcal{L} = \xentropy(\vec{y},\vec{\hat{y}})
\end{equation}

\paragraph{Comparision with \ent.} While our model is largely inspired by Recurrent Entity Networks (\ent[s]: \newcite{Mikael+:2017}), it differs in three main respects. 
First, we explicitly model the delay of activation of the update gates $g^{j}$ with the $\gru$ in \equref[s]{equ:gate2} and (\ref{equ:gru}) as opposed to making $\vec[j]{h}_{i}$ implicitly assume the same responsibility in \ent[s]. Admittedly, for \ent[s] on \babi and \cbt, given the coarse-grained nature and the difference in the granularity of inputs (sentences vs.\ words), the demand for modelling delayed memory update is less obvious. With this delayed gate activation mechanism, we essentially decouple the duty of capturing transitions of activations between steps from the task of entity state tracking. That is, $\vec[j]{h}_{t}$ is now dedicated to keeping track of the state of the $j$-th entity only and released from the burden of monitoring the activation of the update gate. 
Second, tailoring to the task of TABSA, we incorporate not only the target $\vec{t}$ but also the aspect $\vec{a}$ when trying to determine the attention in the $\softmax$ function. Third, the proposed model is bi-directional.

% Local Variables:
% mode: latex
% TeX-master: "naaclhlt2018"
% End:

%% file: ent-small.tex
\begin{tikzpicture}

\node[align=center,minimum size=0.5cm] at (0.0, 0.0) (wt) {\footnotesize  $\vec{w}_{i}$};
\node[align=center] [above left = 0.2 and 2 of wt] (ht_1) {};
\node[align=center] [above = 1.65 of ht_1] (kj) {};
\node[align=center] [above = 0.5 of kj] (dt_1) {};

% phi
\node[draw,align=center,minimum size=0.5cm] [above left = 1.2 and 0.3 of wt] (phi) {\footnotesize $\phi$};

% GRU
\node[draw,align=center,minimum size=0.5cm] [right = 0.98 of dt_1] (gru) {\footnotesize $\gru$};

%\draw [rounded corners,->,>=stealth] (wt.north) |- node [midway,above,shift={(10mm,-1mm)}] (u2m1_label) {\footnotesize Inner product} ($(ht_1.west) + (0.5,0.0)$);

% k path
\draw [rounded corners,->,>=stealth,name path=k_path] (kj.east) -- node [at start,above,shift={(-0.7,0)},anchor=south west] (kj_label) {\footnotesize key $\vec[j]{k}$} ($(kj.east) + (5,0.0)$);

% d path
\draw [rounded corners,->,>=stealth,name path=d_path] (dt_1.east) -- node [at start,above,shift={(-0.7,0)},anchor=south west] (dt_1_label) {\footnotesize delay $\vec[j]{d}_{i-1}$} (gru.west);

% wt -- phi
\draw [rounded corners,->,>=stealth,name path=w_path] (wt.north) -- ($(wt.north) + (0,0.8)$) -| (phi.south);

% ht_1 -- phi
\draw [rounded corners,->,>=stealth] (ht_1.east) -- ($(ht_1.east) + (0.7,0.0)$) |- ($(phi.west) + (0.0, -0.1)$);

% kj -- phi
\draw [rounded corners,->,>=stealth] (kj.east) -- ($(kj.east) + (0.7,0.0)$) |- ($(phi.west) + (0.0, 0.1)$);

% phi -- gru
\draw [rounded corners,->,>=stealth] (phi.north) -- node [at start,right,shift={(0,-0.2)},anchor=south west,text width=1] (ht_tilde_label) {\footnotesize $\vectilde[j]{h}_i$} (gru.south);

% dt
\draw [rounded corners,->,>=stealth] (gru.east) -- node [at start,above,shift={(0,0)},anchor=south west] (dt_label) {\footnotesize $\vec[j]{d}_{i}$} ($(dt_1.east) + (5.0, 0.0)$);

% sigma
\node[draw,align=center,minimum size=0.5cm] [right = 1.3 of phi] (sigma) {\footnotesize $\sigma$};

% kj * wt
\node[draw,circle,align=center,inner sep=1.5pt] [above = 1.28 of wt] (kjwt) {\tiny $L$};

% wt -- kjwt
\draw [rounded corners,->,>=stealth] (wt.north) -- (kjwt.south);

% kj -- kjwt
\draw [rounded corners,->,>=stealth] (kj.east) -| (kjwt.north);

% kjwt -- sigma
\draw [rounded corners,->,>=stealth] (kjwt.east) -- (sigma.west);

% ht_1 * wt
\node[draw,circle,align=center,inner sep=1.5pt] [below = 0.24 of sigma] (ht_1wt) {\tiny $C$};

% wt -- ht_1wt
\draw [rounded corners,->,>=stealth] (wt.north) |- (ht_1wt.west);

% ht_1 -- ht_1wt
\draw [rounded corners,->,>=stealth] (ht_1.east) -| (ht_1wt.south);

% ht_1wt -- sigma
\draw [rounded corners,->,>=stealth] (ht_1wt.north) -- (sigma.south);

% dt -- sigma
\draw [rounded corners,->,>=stealth] (gru.east) -| (sigma.north);

% g
\node[draw,circle,align=center,inner sep=1.5pt] [right = 0.8 of sigma] (g) {\footnotesize $\odot$};

% sigma -- g
\draw [rounded corners,->,>=stealth] (sigma.east) -- node [at start,below,shift={(-0.1,0.8)},anchor=north west,text width=1] (g_label1) {\footnotesize update} node [at start,below,shift={(-0.1,0.4)},anchor=north west,text width=1] (g_label1) {\footnotesize gate} node [at start,below,shift={(-0.1,0.1)},anchor=north west,text width=1] (g_label1) {\footnotesize $g_i^{j}$} (g.west);

% ht -- g
\draw [rounded corners,->,>=stealth] (phi.north) -- ($(phi.north) + (0,0.8)$) -| (g.north);

% ht
\node[draw,circle,align=center,inner sep=0pt] [right = 4.387 of ht_1] (ht) {\footnotesize $+$};

% h path
\draw [rounded corners,->,>=stealth,name path=h_path] (ht_1.east) -- node [at start,below,shift={(-0.7,0)},anchor=north west] (ht_1_label) {\footnotesize memory $\vec[j]{h}_{i-1}$} (ht.west);

% g -- ht
\draw [rounded corners,->,>=stealth] (g.south) -- (ht.north);

\draw [rounded corners,->,>=stealth] (ht.east) -- node [at start,above,shift={(0.0,0.0)},anchor=south west] (ht_label) {\footnotesize $\vec[j]{h}_i$} ($(ht_1.east) + (5,0)$);

\end{tikzpicture}

%% file: exp.tex
\section{Experiments}

\begin{table*}
\center
\begin{tabular}{lcccccc}
\toprule
\multirow{2}{*}{Model} & \multicolumn{3}{c}{Aspect} & & \multicolumn{2}{c}{Sentiment} \\
\cmidrule{2-4}
\cmidrule{6-7}
 & Acc. & $F_1$ & AUC && Acc. & AUC \\
\midrule
\system{LR} \cite{Saeidi+:2016}         &  --- & 39.3 & 92.4 && 87.5 & 90.5 \\
\system{LSTM-Final} \cite{Saeidi+:2016} &  --- & 68.9 & 89.8 && 82.0 & 85.4 \\
\system{LSTM-Loc} \cite{Saeidi+:2016}   &  --- & 69.3 & 89.7 && 81.9 & 83.9 \\
\system{LSTM+TA+SA} \cite{Ma+:2018}     & 66.4 & 76.7 &  --- && 86.8 & --- \\
\system{SenticLSTM} \cite{Ma+:2018}     & 67.4 & 78.2 &  --- && 89.3 & --- \\
\ent[$\dagger$]                         & 66.3 & 69.8 & 89.5 && 87.6 & 89.7 \\
Our model$\dagger$                      & \textbf{73.5} & \textbf{78.5}  & \textbf{94.4} && \textbf{91.0} & \textbf{94.8} \\
\bottomrule
\end{tabular}
\caption{Performance on \sentihood. We take the results reported in \newcite{Saeidi+:2016} and \newcite{Ma+:2018}, resp; \textbf{Bold} = best performance; ``---'' = not reported; $\dagger$ = average performance over 5 runs.}
\label{tbl:performance}
\end{table*}

\subsection{Experimental Setup}
\paragraph{Dataset.}
To test the effectiveness of our model, we use \sentihood, a dataset constructed by \newcite{Saeidi+:2016} for the purpose of detecting aspects and identifying sentiments for each target--aspect pair, consisting of $5,215$ sentences, $3,862$ of which contain a single target, and the remainder multiple targets. Each sentence is annotated with a list of tuples $\{(t, a, y)\}$ with each identifying the sentiment polarity $y$ towards a specific aspect $a$ of a given target $t$ in $s$. Ultimately, given a sentence $s$, we are interested in both detecting the mention of an aspect $a$ for target $t$ (a label other than \textit{none}), and also identifying the specific sentiment $y$ w.r.t.\ the target--aspect pair. A detailed description of the task is presented in \secref{sec:model}.

\paragraph{Model configuration.}
We initialise our model with \system{GloVe} (300-D, trained on 42B tokens, 1.9M vocab, not updated during training: \newcite{Pennington+:2014})
\footnote{\url{http://nlp.stanford.edu/data/glove.42B.300d.zip}} 
and pre-process the corpus with tokenisation using \system{NLTK} \cite{Bird+:2009} and case folding. 
Training is carried out over $800$ epochs with the FTRL optimiser \cite{Mcmahan+:2013} and a batch size of $128$ and learning rate of $0.05$. We use the following hyper-parameters for weight matrices in both directions: $\mat{R} \in \R^{300\times3}$, $\mat{H}$, $\mat{U}$, $\mat{V}$, $\mat{W}$ are all matrices of size $\R^{300\times300}$, $\vec{v} \in \R^{300}$, and hidden size of the $\gru$ in \equref{equ:gru} is $300$. Dropout is applied to the output of $\phi$ in the final classifier (\equref{equ:classifier}) with a rate of $0.2$. Moreover, we employ the technique introduced by \newcite{Gal+:2016} where the same dropout mask is applied to the input $\vec{w}_{i}$ at every step with a rate of $0.2$. Lastly, to curb overfitting, we regularise the last layer (\equref{equ:classifier}) with an $L_2$ penalty on its weights: $\lambda\|\mat{R}\|$ where $\lambda = 0.001$.

We empirically set the number of memory chains to 6, with the keys of two of them set to the same embeddings as the target words \textit{LOC1} and \textit{LOC2}, resp., and the other 4 chains with free key embeddings which are updated during training, and therefore free to capture any entities.\footnote{In line with the findings of \newcite{Mikael+:2017} that tying key vectors damages model performance, we observed similar performance deterioration when using tied keys only. While we also experimented with various configurations (all tied vs.\ all free), this hybrid setup results in the best performance on the validation set.}

Consistent with \newcite{Saeidi+:2016}, we tackle the data unbalanced problem (\textit{none} $\gg$ \textit{positive} $+$ \textit{negative}) by sampling the same number of training instances within a batch randomly from each class.

\paragraph{Evaluation.}
We benchmark against baseline systems presented in the works of \newcite{Saeidi+:2016} and \newcite{Ma+:2018}: (1) \system{LR}: a logistic regression classifier with $n$-gram and POS tag features; (2) \system{LSTM-Final}: a biLSTM taking the final states as representations; (3) \system{LSTM-Loc}: a biLSTM taking the states at the location where target $t$ is mentioned as representations; (4) \system{LSTM+TA+SA}: a biLSTM equipped with complex target and sentence-level attention mechanisms; (5) \system{SenticLSTM}: an improved version of (4) incorporating the \system{SenticNet} external knowledge base \cite{Cambria+:2016}. We additionally implement a bi-directional \ent with the same hyper-parameter settings and \system{GloVe} embeddings as our model \cite{Mikael+:2017}.

In terms of evaluation, we adopt the standard 70/10/20 train/validation/test split, and report the test performance corresponding to the model with the best validation score. Following \newcite{Saeidi+:2016}, we 
consider the top 4 aspects only (\aspect{general}, \aspect{price}, \aspect{transit-location}, and \aspect{safety}) and 
employ the following evaluation metrics: macro-average $F_1$ and AUC for aspect detection ignoring the \textit{none} class, and accuracy and macro-average AUC for sentiment classification. Following \newcite{Ma+:2018}, we also report strict accuracy for aspect detection, as the fraction of sentences where all aspects are detected correctly.

\subsection{Results}

The experimental results are presented in \tabref{tbl:performance}.

\paragraph{State-of-the-art results.} Our model achieves state-of-the-art results for both aspect detection and sentiment classification. 
It is impressive that the proposed model, equipped only with domain-independent general-purpose \system{GloVe} embeddings, outperforms \system{SenticLSTM}, an approach heavily reliant on external knowledge bases and domain-specific embeddings.

\paragraph{\ent vs.\ our model.} We see consistent performance gains for our model in both aspect detection and sentiment classification, compared to \ent, esp.\ for aspect detection, underlining the benefit of delayed update gate activation.

\begin{figure}
\begin{center}
\resizebox{\columnwidth}{!}{
\input{analysis.tex}
}
\end{center}
\caption{\label{fig:analysis} Example of the gate value $g_{t}$ averaged across memory chains, forward and backward, in $\ent$ vs.\ our model.}
\vspace{-30pt}
\end{figure}
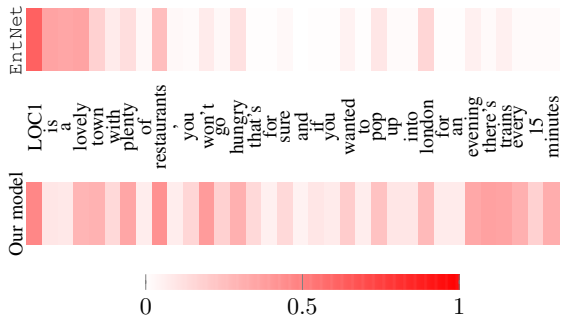

\subsection{Discussion} 

To better understand what the model has learned, we visualise the average gate value $g_t$ in \figref{fig:analysis}, where colour intensity indicates how much memory is updated. Observe that, while updated less by the mention of \textit{LOC1}, our model carries out memory updates upon seeing \textit{lovely town} and \textit{plenty of restaurants}, key phrases associated with aspects such as \aspect{general} and \aspect{dinning}. Perhaps even more importantly, despite the distance between \textit{LOC1} and the final portion of the sentence, our model recognises the relevance to \aspect{transit-location} and keeps the update gates open to track this particular aspect, as opposed to $\ent$ where the last phase is overlooked. The ultimate prediction for the \aspect{transit-location} aspect of \textit{LOC1} is correct with our model (\textit{positive}), but not detected by $\ent$ (\textit{none}), resulting in a false negative. More interestingly, with $\ent$, once distant from a target, it can be frequently observed that the activation rate of $g_t$ tends to drop, a tendency not so apparent with our model.

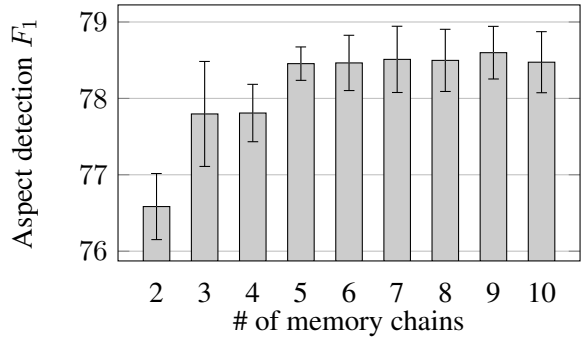
\begin{figure}
\resizebox{\columnwidth}{!}{
\begin{tikzpicture}
      \begin{axis}[
      xlabel=\# of memory chains,
      ylabel=Aspect detection $F_1$,
      width  = \columnwidth,
      height = 5cm,
      major x tick style = transparent,
      ybar=2*\pgflinewidth,
      bar width=10pt,
      ymajorgrids = true,
      symbolic x coords={2,3,4,5,6,7,8,9,10},
      xtick = data,
      scaled y ticks = false
  ]
      \addplot[style={fill=black!20},error bars/.cd, y dir=both, y explicit]
          coordinates {
          (2, 76.584026360880003) += (0,0.43253909338092589) -= (0,0.43253909338092589)
          (3, 77.796377751579996) += (0,0.68717513020859222) -= (0,0.68717513020859222)
          (4, 77.8084018391) += (0,0.3756333219248767) -= (0,0.3756333219248767)
          (5, 78.45484311269999) += (0,0.21866181735551385) -= (0,0.21866181735551385)
          (6, 78.464297252560002) += (0,0.3618687738986423) -= (0,0.3618687738986423)
          (7, 78.510693879320002) += (0,0.43360560353882999) -= (0,0.43360560353882999)
          (8, 78.497480968579993) += (0,0.40690492188256583) -= (0,0.40690492188256583)
          (9, 78.598423984139998) += (0,0.34449250651101299) -= (0,0.34449250651101299)
          (10, 78.473723388939987) += (0,0.39969992768718859) -= (0,0.39969992768718859)};
  \end{axis}
\end{tikzpicture}%
}
\caption{Sensitivity study of model performance to \# of memory chains $n$. Note that we report average performance over 5 runs with standard deviation.}
\label{fig:sensitivity}
\vspace{-30pt}
\end{figure}

In \figref{fig:sensitivity}, we further study the sensitivity of model performance to the number of memory chains $n$ ($2$ of which are constrained to track \lex{LOC1} and \lex{LOC2}, the rest are unconstrained chains). Observe that, when $n < 5$, the model suffers from insufficient capacity (not enough memory chains) to capture the various aspects required by the task, with aspect detection $F_1$ remaining below $78$. In particular, when $n = 2$ (no unconstrained chains), model performance drops substantially to a $F_1$ of $76.6 \pm 0.4$. Once $n \geq 5$, aspect detection $F_1$ increases to around $78$, and is quite stable even with as many as $n = 10$ chains.

% Local Variables:
% mode: latex
% TeX-master: "naaclhlt2018"
% End:

%% file: analysis.tex
\begin{tikzpicture}

\node[align=center,minimum height=0.5cm,rotate=90] at (0.00, 0.0) (w1) {\footnotesize  LOC1};
\node[align=center,minimum height=0.5cm,rotate=90] at (0.25, 0.0) (w2) {\footnotesize  is};
\node[align=center,minimum height=0.5cm,rotate=90] at (0.50, 0.0) (w2) {\footnotesize  a};
\node[align=center,minimum height=0.5cm,rotate=90] at (0.75, 0.0) (w2) {\footnotesize  lovely};
\node[align=center,minimum height=0.5cm,rotate=90] at (1.00, 0.0) (w2) {\footnotesize  town};
\node[align=center,minimum height=0.5cm,rotate=90] at (1.25, 0.0) (w2) {\footnotesize  with};
\node[align=center,minimum height=0.5cm,rotate=90] at (1.50, 0.0) (w2) {\footnotesize  plenty};
\node[align=center,minimum height=0.5cm,rotate=90] at (1.75, 0.0) (w2) {\footnotesize  of};
\node[align=center,minimum height=0.5cm,rotate=90] at (2.00, 0.0) (w2) {\footnotesize  restaurants};
\node[align=center,minimum height=0.5cm,rotate=90] at (2.25, 0.0) (w2) {\footnotesize  ,};
\node[align=center,minimum height=0.5cm,rotate=90] at (2.50, 0.0) (w2) {\footnotesize  you};
\node[align=center,minimum height=0.5cm,rotate=90] at (2.75, 0.0) (w2) {\footnotesize  won't};
\node[align=center,minimum height=0.5cm,rotate=90] at (3.00, 0.0) (w2) {\footnotesize  go};
\node[align=center,minimum height=0.5cm,rotate=90] at (3.25, 0.0) (w2) {\footnotesize  hungry};
\node[align=center,minimum height=0.5cm,rotate=90] at (3.50, 0.0) (w2) {\footnotesize  that's};
\node[align=center,minimum height=0.5cm,rotate=90] at (3.75, 0.0) (w2) {\footnotesize  for};
\node[align=center,minimum height=0.5cm,rotate=90] at (4.00, 0.0) (w2) {\footnotesize  sure};
\node[align=center,minimum height=0.5cm,rotate=90] at (4.25, 0.0) (w2) {\footnotesize  and};
\node[align=center,minimum height=0.5cm,rotate=90] at (4.50, 0.0) (w2) {\footnotesize  if};
\node[align=center,minimum height=0.5cm,rotate=90] at (4.75, 0.0) (w2) {\footnotesize  you};
\node[align=center,minimum height=0.5cm,rotate=90] at (5.00, 0.0) (w2) {\footnotesize  wanted};
\node[align=center,minimum height=0.5cm,rotate=90] at (5.25, 0.0) (w2) {\footnotesize  to};
\node[align=center,minimum height=0.5cm,rotate=90] at (5.50, 0.0) (w2) {\footnotesize  pop};
\node[align=center,minimum height=0.5cm,rotate=90] at (5.75, 0.0) (w2) {\footnotesize  up};
\node[align=center,minimum height=0.5cm,rotate=90] at (6.00, 0.0) (w2) {\footnotesize  into};
\node[align=center,minimum height=0.5cm,rotate=90] at (6.25, 0.0) (w2) {\footnotesize  london};
\node[align=center,minimum height=0.5cm,rotate=90] at (6.50, 0.0) (w2) {\footnotesize  for};
\node[align=center,minimum height=0.5cm,rotate=90] at (6.75, 0.0) (w2) {\footnotesize  an};
\node[align=center,minimum height=0.5cm,rotate=90] at (7.00, 0.0) (w2) {\footnotesize  evening};
\node[align=center,minimum height=0.5cm,rotate=90] at (7.25, 0.0) (w2) {\footnotesize  there's};
\node[align=center,minimum height=0.5cm,rotate=90] at (7.50, 0.0) (w2) {\footnotesize  trains};
\node[align=center,minimum height=0.5cm,rotate=90] at (7.75, 0.0) (w2) {\footnotesize  every};
\node[align=center,minimum height=0.5cm,rotate=90] at (8.00, 0.0) (w2) {\footnotesize  15};
\node[align=center,minimum height=0.5cm,rotate=90] at (8.25, 0.0) (w2) {\footnotesize  minutes};

\node[align=center,minimum width=1.0cm,rotate=90,fill=red!62] at (0.00, 1.4) (w1) {};
\node[align=center,minimum width=1.0cm,rotate=90,fill=red!36] at (0.25, 1.4) (w1) {};
\node[align=center,minimum width=1.0cm,rotate=90,fill=red!35] at (0.50, 1.4) (w1) {};
\node[align=center,minimum width=1.0cm,rotate=90,fill=red!36] at (0.75, 1.4) (w1) {};
\node[align=center,minimum width=1.0cm,rotate=90,fill=red!18] at (1.00, 1.4) (w1) {};
\node[align=center,minimum width=1.0cm,rotate=90,fill=red!08] at (1.25, 1.4) (w1) {};
\node[align=center,minimum width=1.0cm,rotate=90,fill=red!13] at (1.50, 1.4) (w1) {};
\node[align=center,minimum width=1.0cm,rotate=90,fill=red!03] at (1.75, 1.4) (w1) {};
\node[align=center,minimum width=1.0cm,rotate=90,fill=red!26] at (2.00, 1.4) (w1) {};
\node[align=center,minimum width=1.0cm,rotate=90,fill=red!03] at (2.25, 1.4) (w1) {};
\node[align=center,minimum width=1.0cm,rotate=90,fill=red!02] at (2.50, 1.4) (w1) {};
\node[align=center,minimum width=1.0cm,rotate=90,fill=red!08] at (2.75, 1.4) (w1) {};
\node[align=center,minimum width=1.0cm,rotate=90,fill=red!03] at (3.00, 1.4) (w1) {};
\node[align=center,minimum width=1.0cm,rotate=90,fill=red!12] at (3.25, 1.4) (w1) {};
\node[align=center,minimum width=1.0cm,rotate=90,fill=red!01] at (3.50, 1.4) (w1) {};
\node[align=center,minimum width=1.0cm,rotate=90,fill=red!01] at (3.75, 1.4) (w1) {};
\node[align=center,minimum width=1.0cm,rotate=90,fill=red!03] at (4.00, 1.4) (w1) {};
\node[align=center,minimum width=1.0cm,rotate=90,fill=red!00] at (4.25, 1.4) (w1) {};
\node[align=center,minimum width=1.0cm,rotate=90,fill=red!01] at (4.50, 1.4) (w1) {};
\node[align=center,minimum width=1.0cm,rotate=90,fill=red!01] at (4.75, 1.4) (w1) {};
\node[align=center,minimum width=1.0cm,rotate=90,fill=red!05] at (5.00, 1.4) (w1) {};
\node[align=center,minimum width=1.0cm,rotate=90,fill=red!01] at (5.25, 1.4) (w1) {};
\node[align=center,minimum width=1.0cm,rotate=90,fill=red!10] at (5.50, 1.4) (w1) {};
\node[align=center,minimum width=1.0cm,rotate=90,fill=red!02] at (5.75, 1.4) (w1) {};
\node[align=center,minimum width=1.0cm,rotate=90,fill=red!02] at (6.00, 1.4) (w1) {};
\node[align=center,minimum width=1.0cm,rotate=90,fill=red!16] at (6.25, 1.4) (w1) {};
\node[align=center,minimum width=1.0cm,rotate=90,fill=red!01] at (6.50, 1.4) (w1) {};
\node[align=center,minimum width=1.0cm,rotate=90,fill=red!01] at (6.75, 1.4) (w1) {};
\node[align=center,minimum width=1.0cm,rotate=90,fill=red!06] at (7.00, 1.4) (w1) {};
\node[align=center,minimum width=1.0cm,rotate=90,fill=red!02] at (7.25, 1.4) (w1) {};
\node[align=center,minimum width=1.0cm,rotate=90,fill=red!06] at (7.50, 1.4) (w1) {};
\node[align=center,minimum width=1.0cm,rotate=90,fill=red!02] at (7.75, 1.4) (w1) {};
\node[align=center,minimum width=1.0cm,rotate=90,fill=red!02] at (8.00, 1.4) (w1) {};
\node[align=center,minimum width=1.0cm,rotate=90,fill=red!02] at (8.25, 1.4) (w1) {};

\node[align=center,minimum width=1.0cm,rotate=90,fill=red!47] at (0.00, -1.4) (w1) {};
\node[align=center,minimum width=1.0cm,rotate=90,fill=red!10] at (0.25, -1.4) (w1) {};
\node[align=center,minimum width=1.0cm,rotate=90,fill=red!09] at (0.50, -1.4) (w1) {};
\node[align=center,minimum width=1.0cm,rotate=90,fill=red!29] at (0.75, -1.4) (w1) {};
\node[align=center,minimum width=1.0cm,rotate=90,fill=red!30] at (1.00, -1.4) (w1) {};
\node[align=center,minimum width=1.0cm,rotate=90,fill=red!15] at (1.25, -1.4) (w1) {};
\node[align=center,minimum width=1.0cm,rotate=90,fill=red!35] at (1.50, -1.4) (w1) {};
\node[align=center,minimum width=1.0cm,rotate=90,fill=red!07] at (1.75, -1.4) (w1) {};
\node[align=center,minimum width=1.0cm,rotate=90,fill=red!43] at (2.00, -1.4) (w1) {};
\node[align=center,minimum width=1.0cm,rotate=90,fill=red!07] at (2.25, -1.4) (w1) {};
\node[align=center,minimum width=1.0cm,rotate=90,fill=red!16] at (2.50, -1.4) (w1) {};
\node[align=center,minimum width=1.0cm,rotate=90,fill=red!39] at (2.75, -1.4) (w1) {};
\node[align=center,minimum width=1.0cm,rotate=90,fill=red!18] at (3.00, -1.4) (w1) {};
\node[align=center,minimum width=1.0cm,rotate=90,fill=red!31] at (3.25, -1.4) (w1) {};
\node[align=center,minimum width=1.0cm,rotate=90,fill=red!15] at (3.50, -1.4) (w1) {};
\node[align=center,minimum width=1.0cm,rotate=90,fill=red!06] at (3.75, -1.4) (w1) {};
\node[align=center,minimum width=1.0cm,rotate=90,fill=red!15] at (4.00, -1.4) (w1) {};
\node[align=center,minimum width=1.0cm,rotate=90,fill=red!05] at (4.25, -1.4) (w1) {};
\node[align=center,minimum width=1.0cm,rotate=90,fill=red!10] at (4.50, -1.4) (w1) {};
\node[align=center,minimum width=1.0cm,rotate=90,fill=red!08] at (4.75, -1.4) (w1) {};
\node[align=center,minimum width=1.0cm,rotate=90,fill=red!20] at (5.00, -1.4) (w1) {};
\node[align=center,minimum width=1.0cm,rotate=90,fill=red!07] at (5.25, -1.4) (w1) {};
\node[align=center,minimum width=1.0cm,rotate=90,fill=red!24] at (5.50, -1.4) (w1) {};
\node[align=center,minimum width=1.0cm,rotate=90,fill=red!10] at (5.75, -1.4) (w1) {};
\node[align=center,minimum width=1.0cm,rotate=90,fill=red!10] at (6.00, -1.4) (w1) {};
\node[align=center,minimum width=1.0cm,rotate=90,fill=red!27] at (6.25, -1.4) (w1) {};
\node[align=center,minimum width=1.0cm,rotate=90,fill=red!08] at (6.50, -1.4) (w1) {};
\node[align=center,minimum width=1.0cm,rotate=90,fill=red!08] at (6.75, -1.4) (w1) {};
\node[align=center,minimum width=1.0cm,rotate=90,fill=red!34] at (7.00, -1.4) (w1) {};
\node[align=center,minimum width=1.0cm,rotate=90,fill=red!37] at (7.25, -1.4) (w1) {};
\node[align=center,minimum width=1.0cm,rotate=90,fill=red!36] at (7.50, -1.4) (w1) {};
\node[align=center,minimum width=1.0cm,rotate=90,fill=red!31] at (7.75, -1.4) (w1) {};
\node[align=center,minimum width=1.0cm,rotate=90,fill=red!18] at (8.00, -1.4) (w1) {};
\node[align=center,minimum width=1.0cm,rotate=90,fill=red!32] at (8.25, -1.4) (w1) {};

\node[align=center,minimum height=0.5cm,rotate=90] at (-0.30, 1.4) (w1) {\footnotesize  $\ent$};
\node[align=center,minimum height=0.5cm,rotate=90] at (-0.30, -1.4) (w1) {\footnotesize  Our model};

\begin{axis}[
    hide axis,
    scale only axis,
    height=0pt,
    width=0pt,
    at={(0.23\linewidth,-0.27\linewidth)},
    colormap={whiteblue}{color=(white) color=(red)},
    colorbar horizontal,
    point meta min=0.0,
    point meta max=1.0,
    colorbar style={
        width=5cm,
        height=0.25cm,
        xtick={0.0,0.5,1.0},
        axis line style={draw=none},
    }],
    \addplot [draw=none] coordinates {(0.0,0.0)};
\end{axis}

\end{tikzpicture}

%% file: conc.tex
\section{Conclusion}

In this paper, we have proposed a model which is capable of dynamically tracking entities with a delayed memory update mechanism, and demonstrated the effectiveness of the method over the task of targeted aspect-based sentiment analysis. 

% Local Variables:
% mode: latex
% TeX-master: "naaclhlt2018"
% End: